\documentclass[lettersize,journal]{IEEEtran}
\usepackage{amsmath,amsfonts}
\usepackage{algorithm}
\usepackage{algpseudocode}
\usepackage{algorithmicx}
\usepackage{array}
\usepackage{textcomp}
\usepackage{stfloats}
\usepackage{url}
\usepackage{verbatim}
\usepackage{caption}
\usepackage{graphicx}
\usepackage{subfig}
\usepackage[symbol]{footmisc}
\usepackage{hyperref}
\def\BibTeX{{\rm B\kern-.05em{\sc i\kern-.025em b}\kern-.08em
    T\kern-.1667em\lower.7ex\hbox{E}\kern-.125emX}}
\usepackage{balance}

\begin{document}
\title{Strangeness-driven Exploration in Multi-Agent Reinforcement Learning}

\author{
    \IEEEauthorblockN{
        Ju-Bong~Kim,
        Ho-Bin~Choi, and
        Youn-Hee~Han\IEEEauthorrefmark{1}\thanks{\IEEEauthorrefmark{1} Corresponding author.}}
    \IEEEauthorblockA{
        \\ Department of Computer Science Engineering,
        Korea University of Technology and Education\\
        Email:\{rlawnqhd, chb3350, yhhan\}@koreatech.ac.kr
    }
}

\maketitle

\begin{abstract}
Efficient exploration strategy is one of essential issues in cooperative multi-agent reinforcement learning (MARL) algorithms requiring complex coordination.
In this study, we introduce a new exploration method with the strangeness that can be easily incorporated into any centralized training and decentralized execution (CTDE)-based MARL algorithms.
The strangeness refers to the degree of unfamiliarity of the observations that an agent visits.
In order to give the observation strangeness a global perspective, it is also augmented with the the degree of unfamiliarity of the visited entire state.
The exploration bonus is obtained from the strangeness and the proposed exploration method is not much affected by stochastic transitions commonly observed in MARL tasks.
To prevent a high exploration bonus from making the MARL training insensitive to extrinsic rewards, we also propose a separate action-value function trained by both extrinsic reward and exploration bonus, on which a behavioral policy to generate transitions is designed based.
It makes the CTDE-based MARL algorithms more stable when they are used with an exploration method.
Through a comparative evaluation in didactic examples and the StarCraft Multi-Agent Challenge, we show that the proposed exploration method achieves significant performance improvement in the CTDE-based MARL algorithms.
\end{abstract}

\section{Introduction}\label{sec:introduction}
Cooperative multi-agent reinforcement learning (MARL) is an important tool for solving several real-world problems modeled as multi-agent systems, such as autonomous vehicles \cite{IEEETransactions2012Cao} and swarms of robots \cite{10.5555/3322706.3361995}.
In such systems, agents need to train distributed policies through local action-observation histories owing to partial observability.
In MARL, distributed policies exacerbate the issues of joint action spaces that grow exponentially as the number of agents increases.
These issues are addressed through centralized training and decentralized execution (CTDE) \cite{Oliehoek2008OptimalAA}, where policy optimization at the individual level leads to the optimization of a joint policy.

Agents aim to train their policies that maximize the cumulative reward from an unknown environment.
The more complex the dynamics of the environment, the more the agents need to sufficiently explore to discover effective control policies.
Agents can influence the exploration of other agents when training individual policies.
The agents attempt to make cooperative decisions in response to the strategies of others, however, the distributed policies are often trained in a way that only maximizes the benefits of each agent.
As training progresses, policy changes by counterintuitive and non-stationary characteristics reduce the effectiveness of exploration and lead to failure of optimal joint policy training \cite{DBLP:journals/corr/Hernandez-LealK17}.
For these reasons, effective exploration methods in MARL are essential.

Exploration methods in large state and action spaces have been extensively studied in single-agent reinforcement learning (RL).
In the studies about pseudo-counts \cite{10.5555/3157096.3157262} and information gain \cite{DBLP:journals/corr/HouthooftCDSTA16}, the authors propose methods of measuring the state novelty for every action selection in an environment.
However, it is difficult to directly extend them to MARL.
In RL, the curiosity about the unknown state observed by an agent has been the primary measure of the degree of exploration to achieve a goal.
Most curiosity-driven exploration methods \cite{Stadie2015IncentivizingEI, 10.5555/3305890.3305968} utilize bonus rewards based on the error value generated while predicting the next state from the current state and action pair experienced by an agent.
However, when adopting the curiosity-driven exploration methods to MARL, it should be noted that the next state prediction is challenging owing to the actions of other agents (see Fig. 1).
In other words, it is difficult to use curiosity-driven exploration methods in an environment where multiple policies that determine actions based on the current states are non-stationary.

\begin{figure}[t]
    \centering
    \includegraphics[width=0.9\columnwidth]{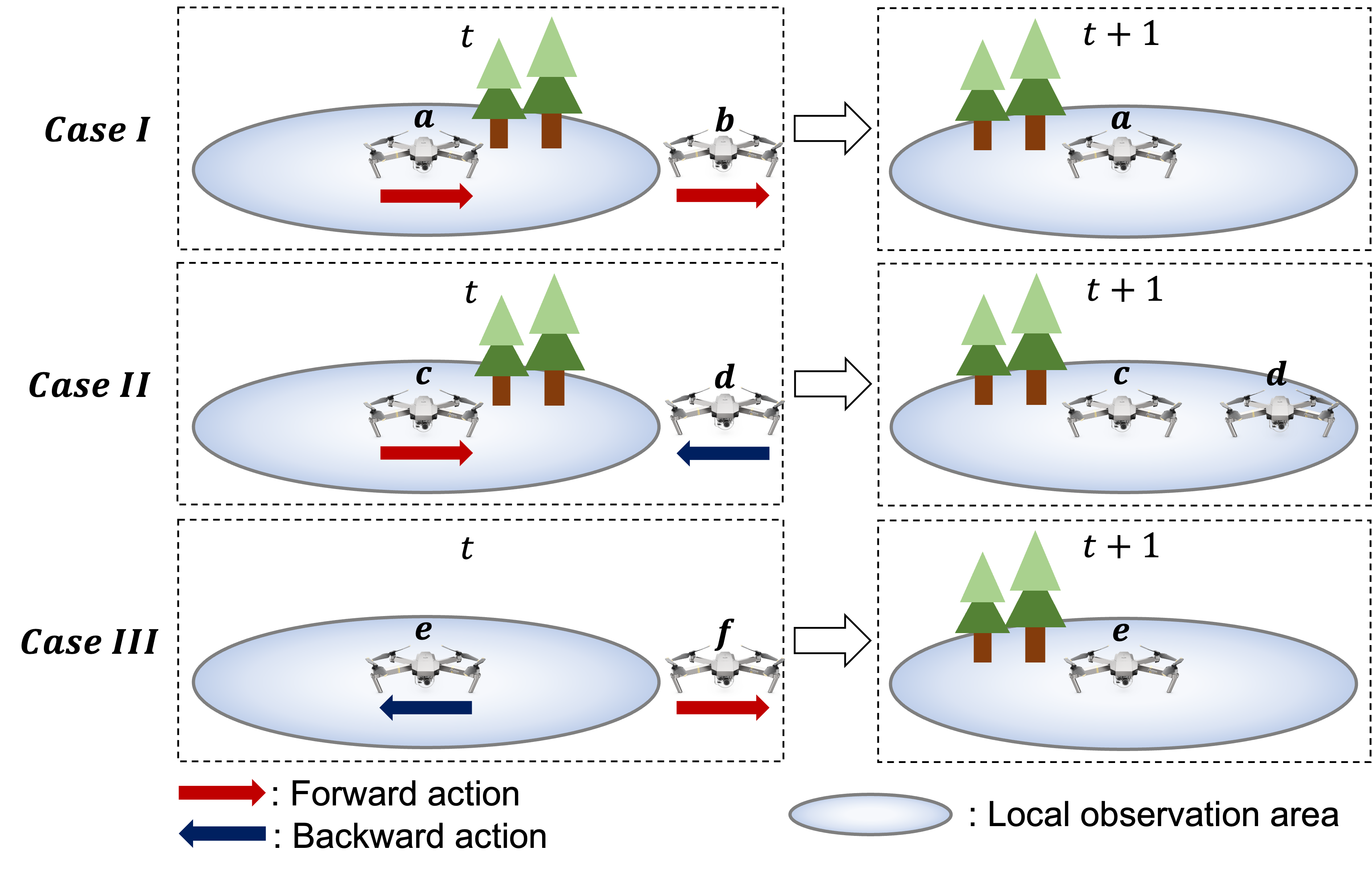} % Reduce the figure size so that it is slightly narrower than the column.
    \caption{Comparative examples of local observation changes in a multi-drone control system. In cases $I$ and $II$, although agents (drones) $\textit{\textbf{a}}$ and $\textit{\textbf{c}}$ have the same local observation and control action, their next local observations differ after the control actions are applied because their neighbor agents $\textit{\textbf{b}}$ and $\textit{\textbf{d}}$ execute different control actions. Conversely, in cases $I$ and $III$, although agents $\textit{\textbf{a}}$ and $\textit{\textbf{e}}$ have different local observation and control action, their next local observations are the same after the control actions are applied. Due to the stochasticity of MARL, as shown in the above cases, it is challenging to predict the next local observation of an agent based on its current local observations and actions in MARL.}
    \label{fig:motiv_new}
\end{figure}

Another exploration method, called random network distillation (RND) \cite{burda2018exploration}, uses a randomly initialized and fixed neural network as a representation network and predicts the embedding of the next state by training a predictor network.
It does not utilize the agents' current state and action pairs to derive the exploration bonuses, so that it may not be significantly affected by the dynamics of the environment.
However, the representation can fail to embed essential state features because the fixed neural network is not trained.
In addition, the transition history over a long period, which may include the {\it strangeness} regarding past sequence of observations, is not used to generate the exploration bonus.

On the other hand, there are other studies of exploration methods that can be directly applied to MARL \cite{iqbal2020coordinated, Mahajan2019MAVENMV, NEURIPS2021_1e8ca836}.
Among them, EMC performs a curiosity-driven exploration by predicting individual action-values based on local action-observation histories.
The action-value function is trained by associating the multiple agents’ current states and actions with the next states.
As shown in from Fig. \ref{fig:motiv_new}, however, such an association is complicated due to the stochasticity of MARL.
Therefore, sometimes, the EMC exploration method fails to effectively train the MARL algorithms.

In this study, we introduce a strangeness-driven exploration method that can be easily incorporated into any CTDE-based MARL algorithms.
To eliminate the direct dependency on the current observation and action, the proposed exploration bonus is generated by the recurrent autoencoder model reconstructing the visited (i.e., next) observation, and the other model predicting the visited (i.e., next) state.
Note that this is interpreted as a problem that produces embeddings of visited observations and states well.
This problem aims to reduce {\it strangeness} by having agents visit observations and states that they have not experienced before.
In other words, the strangeness refers to the degree of strangeness of the observations and states that the agents visit after the current observations, states, and actions have been determined.
Therefore, it is not much affected by forwarding dynamics and stochastic transitions commonly observed in MARL, but only by the strangeness of the visited observations and states.
In this paper, we have the strangeness being high when the next observations and the next states are less experienced, or when the next observations and the next states are visited through less experienced past observation histories.
In addition, we introduce a separate action-value function to make use of the exploration bonus only for efficiently exploring the given environment and to prevent the main MARL training from being significantly affected by the exploration bonus.

We evaluate the proposed exploration method using two didactic examples and the StarCraft Multi-Agent Challenge (SMAC) \cite{DBLP:journals/corr/abs-1902-04043}.
In the didactic example requiring long-term coordination between multiple agents, an MARL algorithm gets only local-optimal solution, whereas the MARL algorithm with the proposed exploration method can get a global-optimal solution successfully.
The results of the comparison study performed on SMAC indicate that the proposed method is superior to other exploration methods.
Moreover, when it was applied to the well-known CTDE-based MARL algorithms, significant performance improvement is observed in the most complex tasks.
To the best of our knowledge, the state-of-the-art results are also achieved for the super hard maps of SAMC.

\section{Preliminaries}

\subsection{Model: Dec-POMDP}
A fully cooperative multi-agent task can be modeled as a decentralized partially observable Markov decision process (Dec-POMDP) \cite{Oliehoek2008OptimalAA}, defined by $\mathcal{G} = \langle\mathcal{S}, \mathcal{U}, \mathcal{P}, \mathcal{Z}, \mathcal{O}, \mathcal{N}, r, \gamma\rangle$.
$\mathcal{S}$ corresponds to the state space in the environment, and $\mathcal{U}$ denotes the action space of agents.
At every time step $t$, agent $i\in \mathcal{N}\equiv \{1, 2, ..., N\}$ selects its action $u^i_t \in \mathcal{U}$, and a joint action $\mathbf{u}_t \equiv [u^i_t]^N_{i=1} \in \mathcal{U}^\mathcal{N}$ is formed with the actions of all agents.
The state transition probability function is defined as $\mathcal{P}(s_{t+1} | s_t,\mathbf{u}_t)\!\!:\mathcal{S} \times \mathcal{U}^\mathcal{N} \times \mathcal{S} \rightarrow [0, 1]$.
All agents receive a joint reward, and its function is defined as $r(s_t,\mathbf{u}_t)\!\!:\mathcal{S} \times \mathcal{U}^\mathcal{N} \rightarrow \mathbb{R}$.
The reward discount factor is $\gamma \in [0, 1)$.
An agent can only observe a local observation $z^i_t \in \mathcal{Z}$ of the state given by the environment through $\mathcal{O}(s_t, i)\!\!:\mathcal{S} \times \mathcal{N} \rightarrow \mathcal{Z}$, and the joint observation $\mathbf{z}_t \equiv [z^i_t]^N_{i=1} \in \mathcal{Z}^\mathcal{N}$ is formed with all the local observations.
The action-observation history for agent $i$ is defined as $\tau^i_t \in \mathcal{T} \equiv (\mathcal{Z} \times \mathcal{U})^{*}$, where $\tau^i_t$ conditions the policy $\pi^{i} (u^i_t | \tau^i_t)\!\!: \mathcal{T} \times \mathcal{U} \rightarrow [0, 1]$.
When the joint history $\boldsymbol{\tau}_t \equiv [\tau^i_t]^N_{i=1}$ is formed with the action-observation histories of all agents, the joint policy $\boldsymbol{\pi} \equiv [\pi^i]^N_{i=1}$ depends on the joint action-value function $Q^{\boldsymbol{\pi}}(\boldsymbol{\tau}_t, \mathbf{u}_t)$ which is defined as:
\begin{align}
    Q^{\boldsymbol{\pi}}(\boldsymbol{\tau}_{t}, \mathbf{u}_t) = \mathbb{E}_{\boldsymbol{\tau}_{t+1:\infty}, \mathbf{u}_{t+1:\infty}} \left[\sum\limits_{k=0}^\infty \gamma^{k}r_{t+k}|\boldsymbol{\tau}_{t}, \mathbf{u}_t\right]
    \label{joint_action_value_function}
\end{align}
and the greedy joint policy is expressed as $\boldsymbol{\pi}(\boldsymbol{\tau}_{t}, \mathbf{u}_{t}) := argmax_{\mathbf{\bar{u}_t}} Q^{\boldsymbol{\pi}}(\boldsymbol{\tau}_{t}, \mathbf{\bar{u}_t})$.
The ultimate goal of cooperative multi-agent is to find the optimal action-value function $Q^{*}$ and the optimal policy $\boldsymbol{\pi}^{*}$.

\subsection{CTDE-based MARL Algorithms}
The CTDE algorithms allow an individual policy $\pi^i$ to access global information in a centralized manner during training, even with limited partial observability and communication constraints.
Based on the CTDE paradigm, many MARL algorithms such as VDN \cite{10.5555/3237383.3238080}, QMIX \cite{pmlr-v80-rashid18a}, Weighted-QMIX \cite{NEURIPS2020_73a427ba}, QTRAN \cite{pmlr-v97-son19a} and QPLEX \cite{wang2021qplex} have been proposed.
They have access to the experiences of joint action-observation histories from a replay memory $\mathcal{D}$ during training, which used for updating the joint policy $\boldsymbol{\pi}(\boldsymbol{\tau}_{t}, \mathbf{u}_{t})$ based on the off-policy learning.
Each individual agent $i$ with policy $\pi^i$ uses only local action-observation history $\tau^i$ to collect experiences, and constructs the policy based on the individual action-value function $Q^{\pi^{i}}(\tau^{i}_{t}, u^{i}_{t})$.

The CTDE algorithms allow the agents to optimize their individual action-value functions through joint action-value function $Q^{\boldsymbol{\pi}}(\boldsymbol{\tau}, \mathbf{u})$ training by minimizing the expected TD-error:
\begin{align}
    \mathcal{L}(\theta) = \mathbb{E}_{\boldsymbol{\tau}, \mathbf{u}, r, \boldsymbol{\tau}^{'} \in \mathcal{D}} \left[y - Q^{\boldsymbol{\pi}}(\boldsymbol{\tau}, \mathbf{u}; \theta) \right]^{2}
    \label{td_error}
\end{align}
where $y = r+\gamma max_{\mathbf{\bar{u}}} Q^{\boldsymbol{\pi}}(\boldsymbol{\tau}', \mathbf{\bar{u}}; \theta^-)$, $\boldsymbol{\tau}'$ is the next joint history, $\theta^-$ is the parameters of the target action-value function, and $\mathcal{D}$ is the replay memory.

\section{Related Works}

\subsection{Curiosity-driven Exploration}

Curiosity-driven exploration aims to minimize the agent's lack of understanding of the environment.
Curiosity is triggered when the agent realizes that it does not have sufficient knowledge of the environment.
In general, the degree of curiosity is inferred by the prediction error of a forward dynamics that predicts the next state based on the current state and action of an agent.
The fact that an agent cannot make accurately predict the forward dynamics for a given current state and action pair means that the agent does not have sufficient knowledge of the dynamics at the state and action pair.
From the prediction error at time $t$, the exploration bonus (i.e., intrinsic reward) $r^{int}_{t}$ is calculated.
The exploration bonus $r^{int}_{t}$ is added to the extrinsic reward $r^{ext}_{t}$ generated from the environment and the mixed reward used for training is denoted by $r_{t} = r^{ext}_{t} + r^{int}_{t}$.
As the predictive performance improves over time, the curiosity (i.e., exploration bonus) gradually diminishes.

A previous study \cite{Stadie2015IncentivizingEI} proposed a method that generates the exploration bonus based on the model of environment dynamics.
Based on the given state encoding function $\sigma$, the predictive model $f_{\theta_{f}}$ parameterized by $\theta_{f}$ predicts $f_{\theta_{f}}\!\!: \sigma(\mathcal{S}) \times \mathcal{A} \rightarrow \sigma(\mathcal{S})$ dynamics where $\mathcal{A}$ is the action space of a single-agent.
Formally, given the state encoding $\sigma(s_t)$ and action $a_t \in \mathcal{A}$, exploration bonus $e(s_t, a_t)$ is defined as:
\begin{align}
    e(s_t, a_t) = \Vert \sigma(s_{t+1}) - f_{\theta_{f}}(\sigma(s), a_t) \Vert^{2}.
    \label{stadie_prediction_error}
\end{align}
The new mixed reward is defined by the addition of the exploration bonus as follows:
\begin{align}
    r_{t} = r^{ext}_{t} + \beta e(s_t, a_t)
    \label{stadie_reward}
\end{align}
where $\beta > 0$ is a scaling factor.
Another curiosity-driven approach, called intrinsic curiosity module (ICM) \cite{pathak2017curiositydriven}, consists of a forward model $f^{fwd}$ and a inverse model $f^{inv}$.
Similar to \cite{Stadie2015IncentivizingEI}, the curiosity in ICM is defined as the error of predicting the result of agent's action based on the current state.
The forward model $f^{fwd}$ receives action $a_t$ and the encoded state $\sigma(s_t)$ at step $t$ and predicts the next encoded state $\sigma(s_{t+1})$.
The inverse model $f^{inv}$ trains the correct state encoding method by correlating the agent's action with the state, and thus predicting the action by using $\sigma(s_t)$ and $\sigma(s_{t+1})$.
In MARL tasks, however, the next local observation is not easy to be predicted due to the stochastic transitions.
Furthermore, the agents sometimes tend to be insensitive to the important extrinsic reward due to excessive intrinsic reward during training time.
In this paper, we propose a new method to avoid these issues.

Additionally, there is another approach called random network distillation (RND) \cite{burda2018exploration}, which uses two neural networks to generate the exploration bonus:
one is a fixed and randomly initialized target network $f$ generating the embedding of a given next state, and the other is a predictor network $f_{\theta_p}$ trained on data collected by the agent.
For a given next state $s_{t+1}$, the predictor network is trained by minimizing the following error:
\begin{align}
    e(s_{t+1}) = \Vert f_{\theta_p}(s_{t+1}) - f(s_{t+1}) \Vert^{2}.
    \label{rnd_prediction_error_end}
\end{align}
From Eq. \eqref{rnd_prediction_error_end}, RND derives the exploration bonus, so that it does not utilize the agents’ current state and action pairs to derive the exploration bonus and it may not be much affected by the dynamics of the environment.
However, the embedding can sometimes fail to include essential features because the fixed embedding neural network is not trained.
In addition, it is difficult to predict the state change over times because the embeddings are generated by exploring the consequences of short-term decisions.
In MARL, changes in patterns of an agent's successive local observations are often stochastic owing to the influence of another agents.
For addressing these issues in MARL, we introduce a similar but new measure, called {\it strangeness}, and suggest a new exploration method that can take into account the strangeness of local observations over long time horizons.

\begin{figure*}[h!]
\centering
\captionsetup{justification=centering}
\includegraphics[width=0.95\textwidth]{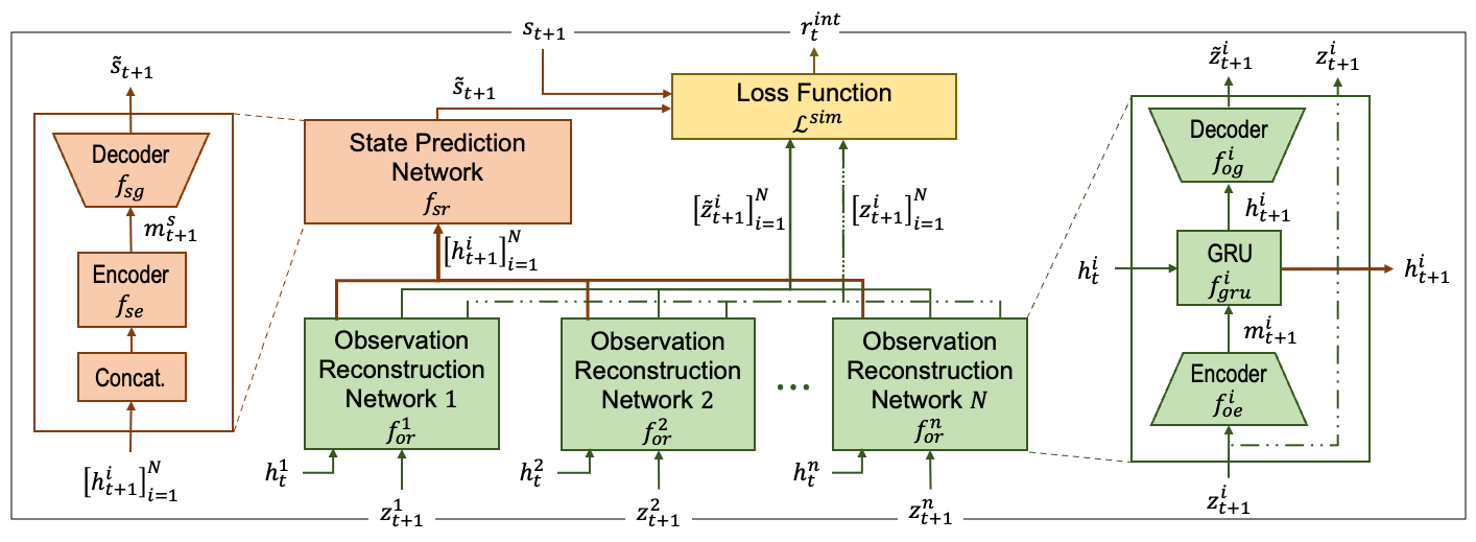} % Reduce the figure size so that it is slightly narrower than the column.
\caption{Overview of \textit{SIM} function $\mathcal{L}^{sim}(\psi) (=r^{int}_{t})$ for $\mathbf{z}_{t+1}$ and $s_{t+1}$ in a transition at time step $t$.}
\label{SIM}
\end{figure*}

\subsection{Exploration Methods in MARL}
Although exploration methods that can be used in single-agent RL have been much studied, there are few easy-to-use effective approaches for fully cooperative MARL.
MULTI \cite{iqbal2020coordinated} and MAVEN \cite{Mahajan2019MAVENMV} introduce a hierarchical structure to facilitate exploration by constructing neural networks independent of each agent's policy.
MULTI presents a framework designed to consider what other agents have explored.
MAVEN introduces a latent space to train the action-value of each agent for hierarchical control.
Based on shared latent variables controlled by hierarchical policies, each agent coordinates its action.
However, MAVEN mainly focuses on overcoming the monotonic limitations of the value factorization rather than on the exploration to discover novel states.

There is another exploration approach named EMC \cite{NEURIPS2021_1e8ca836}.
It performs a curiosity-driven exploration by predicting individual action-values based on local action-observation histories, and utilizes the prediction errors of individual action-values to get the exploration bonus for achieving coordinated exploration.
However, it may make valid exploration delayed because the action-value functions through extrinsic rewards must be much trained in order to perform the coordinated exploration well.
In addition, the action-value function is usually trained by associating the multiple agents’ current states and actions with the next states.
As we explained previously, such association is difficult due to the stochasticity of MARL, and should be avoided for efficient exploration method.

\section{Exploration method with Strangeness} \label{sec:strangeness-driven-exploration}

In this section, we introduce a novel approach that generates the exploration bonus for MARL.
The exploration is to reduce an agent's ignorance by visiting unexplored observations and states.
In this study, the exploration bonus represents the strangeness of local observations and entire state.
The strangeness is meant to be high when:
\begin{enumerate}
\item the visited (i.e., next) observations and states are not much experienced, or
\item the visited (i.e., next) observations and states are reached through past observation histories that are not experienced much.
\end{enumerate}

To measure the strangeness of the visited observation and state quantitatively according to its meaning, we propose the {\it strangeness index module (SIM)}, which is easily calculated at any CTDE-based MARL algorithms.
Without direct dependency on the current observation and action, SIM aims to reconstruct the visited observation through its recurrent autoencoder assuming that previous observations and actions are determined.
In our recurrent autoencoder, GRU \cite{69e088c8129341ac89810907fe6b1bfe} preserves the past experience of each agent, so that the proposed SIM can also measure the strangeness over the transition history.
SIM also predicts the visited entire state through the representation of each agent's next observations, which is produced by the state prediction network.

\subsection{Strangeness Index Module (SIM)}
Fig. \ref{SIM} demonstrates the proposed SIM structure.
For $\mathbf{z}_{t+1}$ and $s_{t+1}$ in a transition $(\mathbf{z}_{t}, s_{t}, \mathbf{u}_{t}, r_t^{ext}, \mathbf{z}_{t+1}, s_{t+1})$ at every time step $t$, the SIM function $\mathcal{L}^{sim}(\psi)$ is the exploration bonus generator parameterized by $\psi$.
It consists of 1) the observation reconstruction networks (recurrent autoencoders) $[f_{or}^{i}]^{N}_{i=1}$ for each agent $i \in (1, ..., N)$ and 2) the state prediction network $f_{sr}$.

For an agent $i$, an observation reconstruction network $f_{or}^{i}$ consists of 1) an encoder $f_{oe}^{i}$ to produce a representation of the next observation, 2) a recurrent history mixer $f_{gru}^{i}$ to embed a hidden state of representations over time via GRU, and 3) a decoder $f_{og}^{i}$ to reconstruct the next observation.
$f_{oe}^i(z^{i}_{t+1})\!:\mathcal{Z} \rightarrow \mathbb{R}^{d}$ takes the next observation $z^{i}_{t+1}$ and produces a $d$-dimensional representation $m^{i}_{t+1}$.
$f_{gru}^i(m^{i}_{t+1}, h^{i}_{t})\!:\mathbb{R}^{d} \times \mathbb{R}^{d} \rightarrow \mathbb{R}^{d}$ takes $m^{i}_{t+1}$ and the hidden state $h^{i}_{t}$ and produces the next hidden state $h^{i}_{t+1}$.
$f_{og}^i(h^{i}_{t+1})\!:\mathbb{R}^{d} \rightarrow \mathcal{Z}$ takes the next hidden state $h^{i}_{t+1}$ and finally produces ${\tilde{z}}^{i}_{t+1}$ (i.e., a prediction of $z^{i}_{t+1}$).

The state prediction network $f_{sr}$ consists of 1) an encoder $f_{se}$ to produce a representation of the state through the representation of observation histories and 2) a decoder $f_{sg}$ to predict the next state.
$f_{se}([h^{i}_{t+1}]^{N}_{i=1})\! :\mathbb{R}^{d} \times \mathcal{N} \rightarrow \mathbb{R}^{d}$ takes the concatenated next hidden states of the agents and produces a $d$-dimensional representation $m^{s}_{t+1}$.
$f_{sg}(m^{s}_{t+1})\!:\mathbb{R}^{d} \rightarrow \mathcal{S}$ takes $m^{s}_{t+1}$ and finally produces ${\tilde{s}}_{t+1}$ (i.e., a prediction of $s_{t+1}$).

The prediction error (i.e.\ strangeness or internal reward) is calculated by loss function $\mathcal{L}^{sim}$ as follows:
\begin{equation}
\begin{aligned}
    \mathcal{L}^{sim}(\psi) &= r^{int}_{t} \\&= \rho \frac{1}{N} \sum\limits_{i=1}^N {\Vert {\tilde{z}}^{i}_{t+1} - z^{i}_{t+1} \Vert^{2}_{2}} + (1 - \rho) \Vert {\tilde{s}}_{t+1} - s_{t+1} \Vert^{2}_{2}
    \label{exp_bonus}
\end{aligned}
\end{equation}
where $\rho$ is the weight balancing the observation reconstruction error and the state prediction error. $r^{int}_{t}$ is used as an exploration bonus (intrinsic reward).
Eventually, The exploration bonus is reduced when agents are well aware of the observations and states they encounter while exploring.
The mixed reward is defined as:
\begin{align}
    r_t = r^{ext}_t + \beta r^{int}_t
    \label{mixed_reward}
\end{align}
where $\beta > 0$ is a scaling factor.

\begin{algorithm}[t!]
\caption{CDTE-based MARL algorithm extended with our exploration method}\label{alg:SIM_algorithm}
\begin{algorithmic}
\State Initialize algorithm parameters: $\alpha, \beta, \gamma$
\State Initialize $Q^{\boldsymbol{\pi}}$ with parameter vectors $\theta$
\State Initialize $Q^{\boldsymbol{\pi}^{exp}}$ with parameter vectors $\omega$
\State Initialize $\mathcal{L}^{sim}$ with parameter vector $\psi$
\State Initialize $\boldsymbol{\pi}$ w.r.t. $Q^{\boldsymbol{\pi}}(\boldsymbol{\cdot}~; \theta)$
\State Initialize $\boldsymbol{\pi}^{exp}$ w.r.t. $Q^{\boldsymbol{\pi}^{exp}}(\boldsymbol{\cdot}~; \omega)$
\State Initialize the replay memory $\mathcal{D} \leftarrow \{\}$
\\
\For {each episode iteration}
    \State Initialize $\boldsymbol{\tau}_0 \neq$ terminal
    \For {each environment step $t$}
        \State Observe $\mathbf{z}_{t}$ and $s_{t}$, and build $\boldsymbol{\tau}_t$
        \State Sample $\mathbf{u}_t$ using $\boldsymbol{\pi}^{exp}$ and $\boldsymbol{\tau}_t$
        \State Execute $\mathbf{u}_t$, observe $\mathbf{z}_{t+1}$ and $s_{t}$, and get $r^{ext}_t$
        \State $\mathcal{D} \leftarrow \mathcal{D} \cup {(\mathbf{z}_t, s_t, \mathbf{u}_t, r^{ext}_t, \mathbf{z}_{t+1}, s_{t+1})}$
    \EndFor
    \For {each gradient step}
        \State Sample a mini-batch $\mathcal{B}$ from $\mathcal{D}$
        \State Calculate $\mathcal{L}(\theta)$ using $\mathcal{B}$ {{\Comment{by Eq. \eqref{goal_model_td_error}}}}
        \State $\theta \leftarrow \theta - \alpha \bigtriangledown\!\mathcal{L}(\theta)$
        \State Calculate $\mathcal{L}^{sim}(\psi)(=r^{int})$ using $\mathcal{B}$ {{\Comment{by Eq. \eqref{exp_bonus}}}}
        \State $\psi \leftarrow \psi - \alpha \bigtriangledown\!\mathcal{L}^{sim}(\psi)$
        \State $r = r^{ext} + r^{int}$
        \State Update $\mathcal{B}$ by replacing $r^{ext}$ with $r$
        \State Calculate $\mathcal{L}^{exp}(\omega)$ using $\mathcal{B}$ {{\Comment{by Eq. \eqref{exploration_model_td_error}}}}
        \State $\omega \leftarrow \omega - \alpha \bigtriangledown\!\mathcal{L}^{exp}(\omega)$
    \EndFor
\EndFor
\end{algorithmic}
\end{algorithm}

\subsection{Exploration Action-Value Function} \label{subsec:exploration_model}

In CTDE-based MARL algorithms using the proposed exploration method, we refer the general joint action-value function as the {\it goal action-value function} $Q^{\boldsymbol{\pi}}(\boldsymbol{\tau}, \mathbf{u}; \theta)$ parameterized by $\theta$.
It is trained like most algorithms by minimizing the following squared TD-error:
\begin{align}
    \mathcal{L}(\theta) = \sum\limits_{\mathcal{B}} \left[ \left(y - Q^{\boldsymbol{\pi}}(\boldsymbol{\tau}, \mathbf{u}; \theta) \right)^{2} \right]
    \label{goal_model_td_error}
\end{align}
where $y = r^{ext}+\gamma max_{\mathbf{u}'} Q^{\boldsymbol{\pi}}(\boldsymbol{\tau}', \mathbf{u}'; \theta^-)$ and $\mathcal{B}$ is the mini-batch from the replay memory $\mathcal{D}$.

Once the agents have explored enough, the SIM's prediction error becomes small.
However, early in training, SIM can often generate the exploration bonus (i.e., intrinsic reward) higher than extrinsic reward.
It is not easy to adjust the balance between extrinsic reward and intrinsic reward only by relying on the scaling factor $\beta$ in Eq. \eqref{mixed_reward}.
The extrinsic reward from a particular state remains the same during the training phase, while the exploration bonus depends on the degree of strangeness for the particular observation and state, or the strangeness over the transition history.
Sometimes, if agents have a high degree of strangeness (i.e., a high exploration reward), it is unluckily possible to make the goal action-value function $Q^{\boldsymbol{\pi}}(\boldsymbol{\tau}_t, \mathbf{u}_t; \theta)$ insensitive to extrinsic rewards for the state.

For this reason, we propose a technique in which the goal action-value function is not directly affected by the exploration bonus.
We build a {\it exploration action-value function} $Q^{\boldsymbol{\pi}^{exp}}(\boldsymbol{\tau}_t, \mathbf{u}_t; \omega)$, on which a behavioral policy $\boldsymbol{\pi}^{exp}$ is designed based, as a joint action-value function parameterized by $\omega$.
The exploration action-value function is trained by minimizing the squared TD-error with the mixed reward in Eq. \eqref{mixed_reward}:
\begin{align}
    \mathcal{L}^{exp}(\omega) = \sum\limits_{\mathcal{B}} \left[ \left(y^{exp} - Q^{\boldsymbol{\pi}^{exp}}(\boldsymbol{\tau}, \mathbf{u}; \omega) \right)^{2} \right]
    \label{exploration_model_td_error}
\end{align}
where
\begin{align}
y^{exp} = r+\gamma Q^{\boldsymbol{\pi}}(\boldsymbol{\tau}', argmax_{\mathbf{u}'} Q^{\boldsymbol{\pi}^{exp}}(\boldsymbol{\tau}', \mathbf{u}' ; \omega) ; \theta^-).
    \label{exploration_model_td_error_y}
\end{align}
In Eq. \eqref{exploration_model_td_error_y}, $r=r^{ext}+\beta r^{int}$ and the target model parameterized by $\theta^-$ corresponds to the goal action-value $Q^{\boldsymbol{\pi}}$.
It means that the $Q^{\boldsymbol{\pi}^{exp}}$ update exploits the exploration bonus as well as the extrinsic reward, but $Q^{\boldsymbol{\pi}}$ is used to calculate the next action-value (i.e., the action-value for the pair of next observation and action) even though $Q^{\boldsymbol{\pi}}$ and $Q^{\boldsymbol{\pi}^{exp}}$ have different purposes.
Since the exploration bonus is usually large and varied at the beginning of training, $Q^{\boldsymbol{\pi}^{exp}}$ rarely converges under the usage of $Q^{\boldsymbol{\pi}}$ for the next action-value, but it might be helpful for exploration.
However, as the training progresses, the exploration bonus will get smaller and $Q^{\boldsymbol{\pi}^{exp}}$ will become more similar to $Q^{\boldsymbol{\pi}}$.
It is the reason why we use the goal action-value $Q^{\boldsymbol{\pi}}$ for the next action-value in Eq. \eqref{exploration_model_td_error_y}.

It is also noted that $y^{exp}$ is calculated based on Double Q-learning algorithm \cite{10.5555/3016100.3016191}, so that
the agents' actions at the next observations are selected based on $Q^{\boldsymbol{\pi}^{exp}}$ when $Q^{\boldsymbol{\pi}}$ is calculated for the next action-value.
For calculating $y^{exp}$, the the standard DQN's max operator would use the same goal action-value both to select and to evaluate the next actions.
It makes it more likely to obtain overestimated value, resulting in overoptimistic goal action-value estimate.
To prevent this, we decouple the action selection from the evaluation in Eq. \eqref{exploration_model_td_error_y}.

The pseudo code of a CTDE-based MARL algorithm using the proposed exploration method is represented in Algorithm \ref{alg:SIM_algorithm}.
It is noted that the experiences generated from the exploration policy $\boldsymbol{\pi}^{exp}$ based on $Q^{\boldsymbol{\pi}^{exp}}(\boldsymbol{\tau}_t, \mathbf{u}_t; \omega)$ are stored in $\mathcal{D}$ and used for training of the exploration and goal action-value functions.
The exploration action-value function is no longer needed when the training is finished.

\section{Experiments}\label{sec:experiments}

In this section, we conduct empirical experiments to answer the following questions:

\begin{figure*}[t]
    \centering
    \subfloat[Perfect trajectory example in 6-step payoff matrix game\label{fig:POM}]
    {
        \centering
        \includegraphics[width=0.48\textwidth]{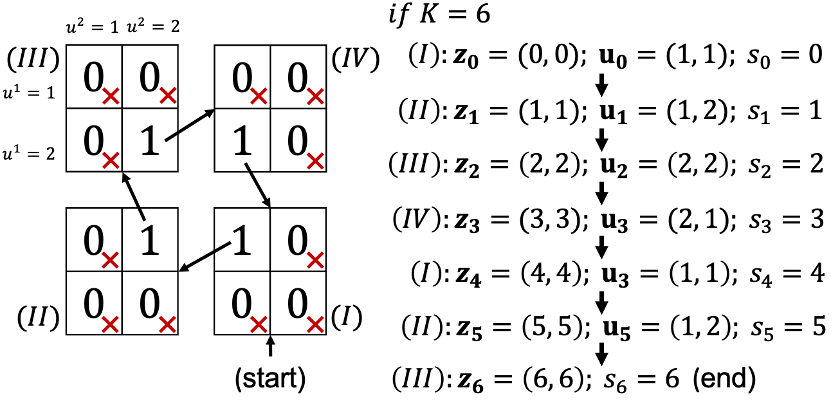}
    }
    \hfill
    \subfloat[Test return and action-value averaged over 20 random initializations on the $128$-step payoff matrix game\label{fig:POM_without_EQ_results}]
    {
        \centering
        \includegraphics[width=0.48\textwidth]{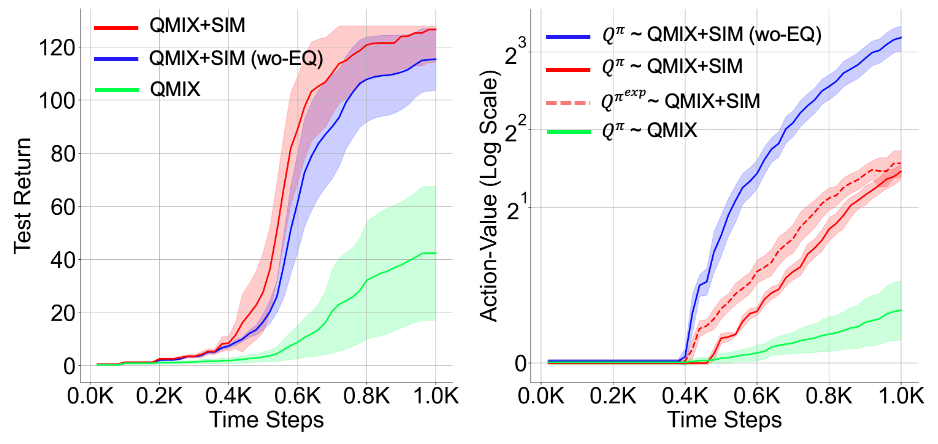}
    }
    \caption{$K$-step payoff matrix game and the ablation experiment in the case of $K=128$.}
    \label{fig:k-step_playoff_matrix}
\end{figure*}

\begin{figure*}[!htb]
    \centering
    \captionsetup{justification=centering}
    \includegraphics[width=2.0\columnwidth]{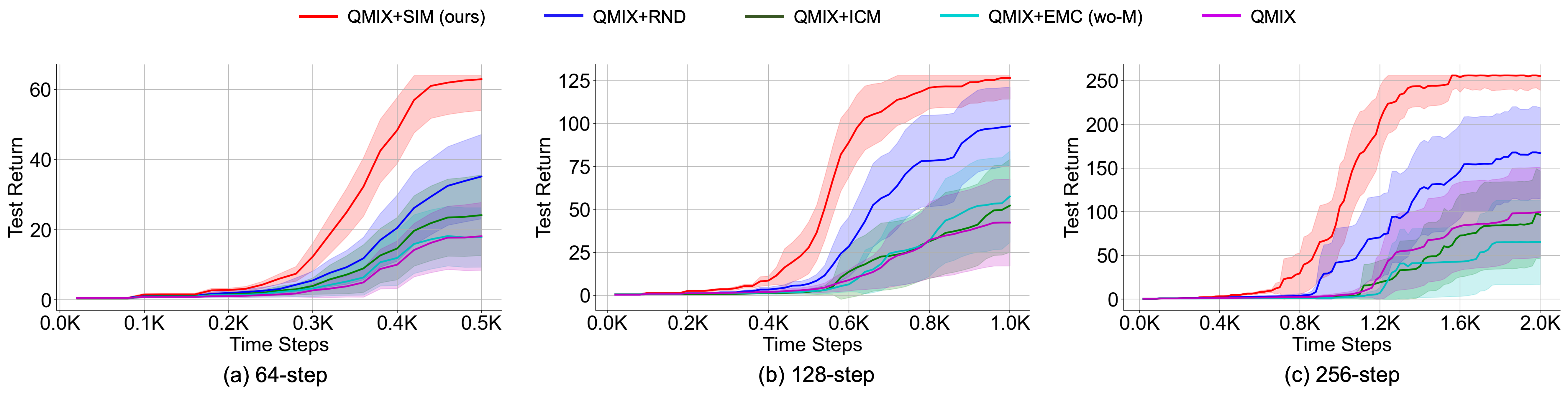} % Reduce the figure size so that it is slightly narrower than the column.
    \caption{Test return averaged over 20 random initializations on the $K$-step payoff matrix game.}
    \label{fig:POM_results}
\end{figure*}

\begin{itemize}
\item Can the proposed exploration method perform effectively in multi-agent tasks that require coordinated decisions over long periods?
\item Can the proposed exploration method be more efficient than the existing well-known exploration methods?
\item Can the proposed exploration method be applied to various MARL algorithms to contribute to more efficient performance improvement?
\end{itemize}
For answering the above questions, first of all, we verify our proposed exploration method, SIM, on two didactic examples.
We then evaluate it on SMAC which provides a diverse set of challenging scenarios with high control complexity.

For comparative analysis, we also extend the two well-known single-agent RL exploration methods, ICM and RND, to support MARL and select them for the baseline exploration methods.
Both ICM and RND produce their prediction errors using local observations of each agent and derive the exploration bonus in the similar manner as in Eq. \eqref{exp_bonus}, but do not consider to predict the visited entire state.
As another baseline exploration method, we also select EMC, a recent exploration method specific to the MARL algorithms; however, the episodic memory function is excluded from EMC to ensure consistency of training methods across the baseline algorithms, and we change its name into EMC (wo-M).
In this section, therefore, a total of four exploration methods, SIM, ICM, RND, and EMC (wo-M), are compared and evaluated.

In this section, all graphs of experimental results represent the mean performance values over multiple runs (each with a different random seed) and their $95\%$ confidence intervals of the mean values.
Details and hyperparameters can be found at \url{https://github.com/JJBong/strangeness_exploration}

\subsection{Didactic Example I: $K$-step Payoff Matrix Game}

Fig. \ref{fig:POM} illustrates the $K$-step payoff matrix game with two agents ($\mathcal{N}=\{1, 2\}$), wherein each row describes the strategy of the first agent and each column describes the strategy of the second agent.
For a game step, the row and column indices are independently picked by the two agents simultaneously (i.e., $|u^1|=|u^2|=2$), and the matrix entry at the intersection of each row and column represents the joint reward taken by the corresponding picked actions.
The observations each agent receives and the entire state equal the number of steps in the game.
The game's steps continue only when the joint reward is $1$, and the maximum step of a game (i.e., an episode) is $K$.
Therefore, an episode terminates 1) when a joint reward is $0$ or 2) when it reaches the $K$ time step.
The experiments are performed with $K=64, 128$ and $256$, and as $K$ increases, more complex coordination are required.

Fig. \ref{fig:POM_without_EQ_results} demonstrates our ablation experiment highlighting the importance of exploration action-value function $Q^{\boldsymbol{\pi}^{exp}}$ when applying the proposed SIM method to QMIX \cite{pmlr-v80-rashid18a}.
QMIX+SIM represents the QMIX algorithm with our exploration method, SIM, whereas QMIX+SIM (wo-EQ) represents the QMIX algorithm with our another method where $Q^{\boldsymbol{\pi}^{exp}}$ is not used and $Q^{\boldsymbol{\pi}}$ is trained with both $r^{ext}$ and $r^{int}$ like the existing curiosity-driven exploration methods.
With regard to the average test return, QMIX+SIM (wo-EQ) almost follows QMIX+SIM.
In QMIX+SIM (wo-EQ), however, $Q^{\boldsymbol{\pi}}$ is heavily affected by the exploration bonus and it may be also overestimated.
In QMIX+SIM, on the other hand, $Q^{\boldsymbol{\pi}}$ is not.
In addition, you can verify that $Q^{\boldsymbol{\pi}^{exp}}$ does not increase much and become similar to $Q^{\boldsymbol{\pi}}$ as the training progresses, since the exploration bonus gets smaller and the goal action-value $Q^{\boldsymbol{\pi}}$ is used for the next action-value in Eq. \eqref{exploration_model_td_error_y}.

In 64, 128, and 256-step payoff matrix games with two agents, as shown in Fig. \ref{fig:POM_results}, QMIX+SIM outperforms the other QMIX extensions, and finally finds the optimal policy after enough training.
As noted in Fig. \ref{fig:motiv_new}, it may not be helpful for an agent to directly use the observation and action of current time step to predict the strangeness of the next observation.
Only two exploration schemes, SIM and RND, do not use the current observation and action to get the exploration bonus.
It may be the reason why both QMIX+SIM and QMIX+RND outperform the others.
Like QMIX+ICM, QMIX+EMC (wo-M) and QMIX, however, QMIX+RND also fails to learn the optimal policy in all settings of payoff matrix games.

\begin{figure*}[t]
    \centering
    \subfloat[An example episode with 4 agents\label{fig:pressure_plate_a}]
    {
        \centering
        \includegraphics[width=0.48\textwidth]{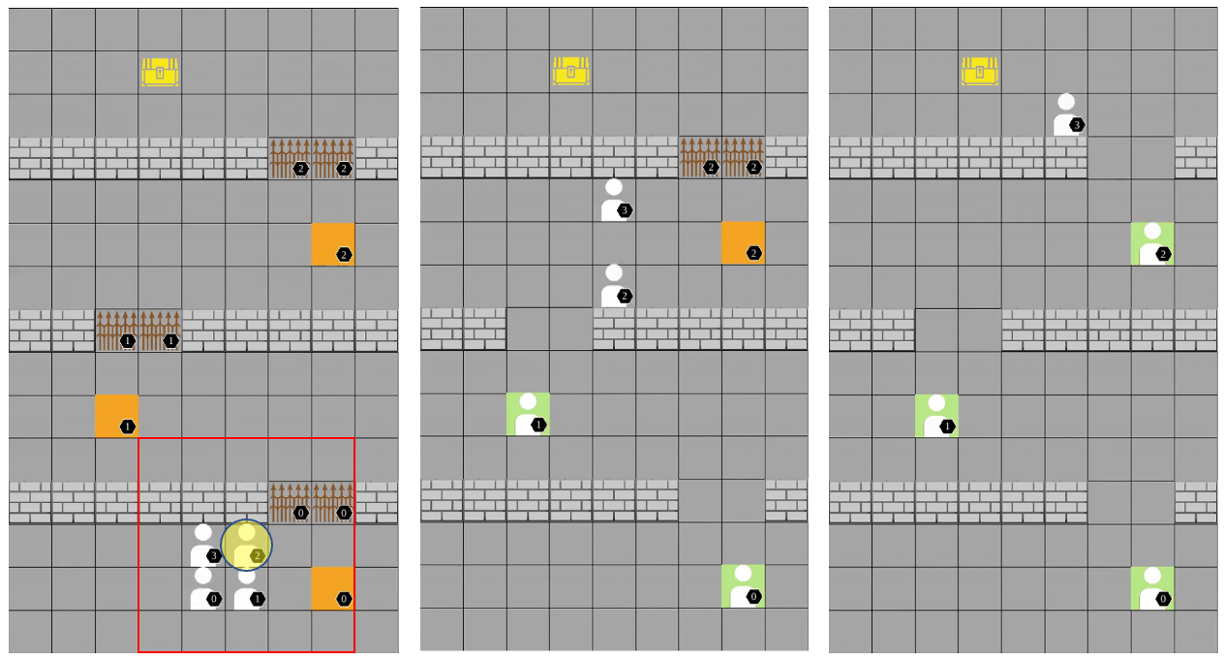}
    }
    \hfill
    \subfloat[Test episode length and exploration bonus over 20 random initializations\label{fig:pressure_plate_b}]
    {
        \centering
        \includegraphics[width=0.48\textwidth]{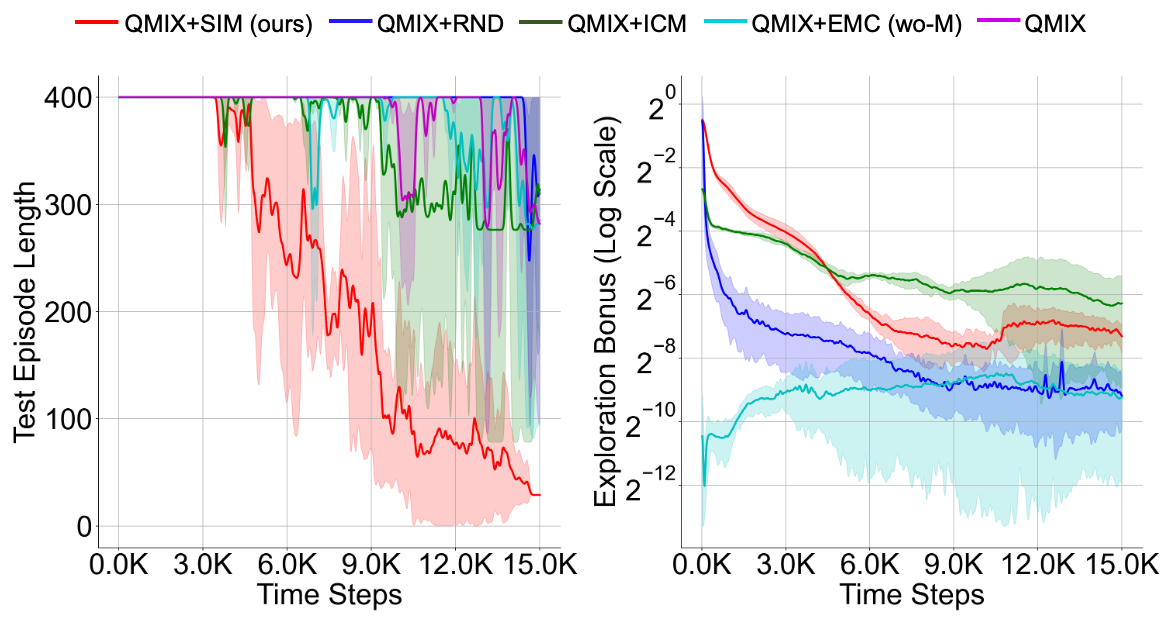}
    }
    \caption{PressurePlate multi-agent environment and its experimental results of our evaluated algorithms}
    \label{fig:pressure_plate}
\end{figure*}

\subsection{Didactic Example II: PressurePlate}\label{subsec:pressureplate}
We also evaluate the proposed exploration method in a grid-based multi-agent open source environment called PressurePlate \cite{pressureplate_github}, where the cooperation between agents is important (see Fig. \ref{fig:pressure_plate_a}).
The environment is made of several 2D grids.
It contains several rooms, and each room has a plate and a closed doorway.
When an episode begins, each agent is assigned a plate that only it can activate.
For the multiple agents to proceed into the next room, an agent must remain behind and stand at the assigned assigned plate.
The task is considered solved when the last room, as depicted with a treasure chest in Fig. \ref{fig:pressure_plate_a}, is reached by the last agent who has been assigned the room's treasure chest.
Each agent has a distance-limited local view ($5 \times 5$) of the environment, as depicted at the leftmost one in Fig. \ref{fig:pressure_plate_a}.
When queried, the environment produces the multiple local views of each which corresponds to each agent's viewing range.
Next, these local views are flattened, concatenated together, and used as the observation vector.
Finally, each agent's coordinates are concatenated to the end of each observation vector.
The action space is discrete and has five options: up, down, left, right, and no-op (do nothing).

Fig. \ref{fig:pressure_plate_b} shows the test episode lengths of evaluated algorithms during training, and also shows the change of the exploration bonuses.
QMIX+SIM is effective in a multi-agent environment with non-stationary characteristics because it's networks predict the next observations and states from histories of each agent after the current observations and actions are determined.
In this environment setting, the agents often fail to train because the joint rewards they receive are sparse and hard to find a way to cooperate in each agent's diverse sequences of transitions.
However, at the beginning of training, QMIX+SIM agents receive relatively high exploration bonuses and are encouraged to explore.
Because of this, it achieves its goal faster than the other algorithms.

\subsection{Exploration Performance Comparison}

\begin{figure*}[!htb]
    \centering
    \captionsetup{justification=centering}
    \includegraphics[width=2.0\columnwidth]{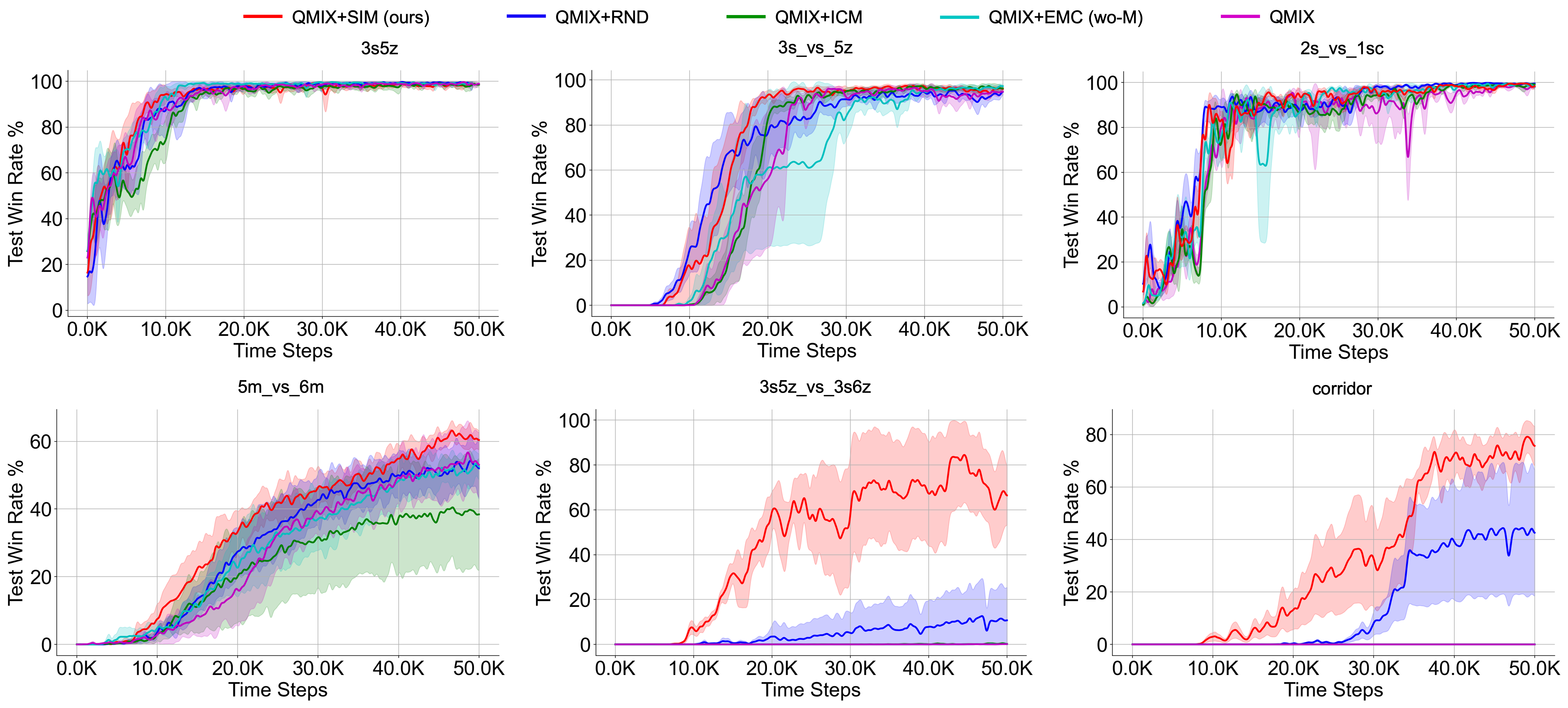} % Reduce the figure size so that it is slightly narrower than the column.
    \caption{Win rates on the six SMAC maps (for exploration performance comparison).}
    \label{fig:QMIX_results}
\end{figure*}

SMAC consists of a set of micro-scenarios in which individual agents cooperate to defeat enemies.
A predetermined number and type of agents battle enemies in each scenario on a given map.
We evaluate the proposed exploration method on six maps with a different purpose and difficulty of scenarios.
Fig. \ref{fig:QMIX_results} shows the rate at which agents trained by our method win a match against enemies controlled by a built-in AI.
Our QMIX extension, QMIX+SIM, outperforms the other QMIX extensions as well as QMIX in all six maps.
For the three maps in the first row of the figure, which do not require significant exploration, the performance of the algorithms does not differ significantly.
However, for the three maps in the second row that require a considerable amount of exploration, it is difficult for all methods to make complex coordination in a short training time
In the three difficult maps, the stochasticity level caused by multiple agents is too high and sparse extrinsic rewards caused by the coordination tend to make the training difficult.
QMIX+ICM and QMIX+EMC (wo-M) fail to train the multiple agents on such difficult maps since they try to associate their current observations and actions with the next observations.
QMIX+RND makes agents become much sensitive to the exploration bonus due to the sparse extrinsic reward.
However, QMIX+SIM allows the multiple agents to effectively make complex coordination within a short period.
To the best of our knowledge, the state-of-the-art results are achieved for the super hard maps, {\it 3s5z\_vs\_3s6z} and {\it corridor}.

\begin{figure*}[!htb]
    \centering
    \captionsetup{justification=centering}
    \includegraphics[width=2.0\columnwidth]{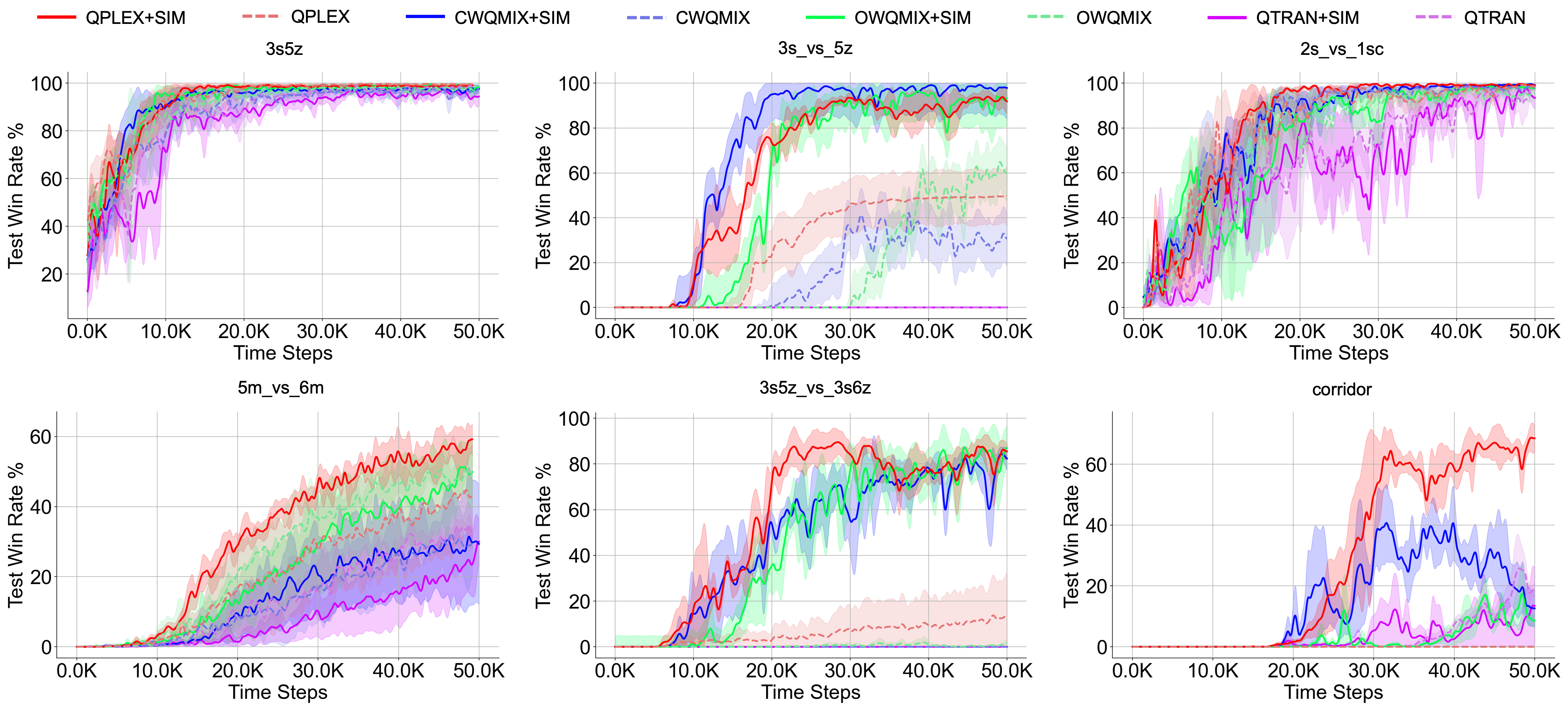} % Reduce the figure size so that it is slightly narrower than the column.
    \caption{Win rates on the six SMAC maps (for applicability test).}
    \label{fig:CTDE_results_2}
\end{figure*}

\subsection{Applicability}
Our exploration method is applicable to any CTDE-based MARL algorithms and can improve performance.
To address this, we apply it to the well-known CTDE-based MARL algorithms, QPLEX, centrally weighted QMIX (CWQMIX), optimistically weighted QMIX (OWQMIX), and QTRAN, and evaluate their performance.
Fig. \ref{fig:CTDE_results_2} shows the results of evaluating the above algorithms on all the six maps of SMAC.
It can be observed that the four algorithms, QPLEX+SIM, CWQMIX+SIM, OWQMIX+SIM, and QTRAN+SIM, extended with the proposed exploration method are superior to the original algorithms in most maps of SMAC.
Most CTDE-based MARL algorithms tend to spend a significant amount of time performing exploration on complex tasks.
Our exploration method can significantly reduce this wasted time and improve the training performance.

\section{Conclusion and Future Work}
In this study, we propose a new exploration method, SIM, where the strangeness is defined as a new measure of exploration bonus.
With a past observation history of an individual agent, it generates an exploration bonus by calculating the reconstruction error for the visited observation via recurrent autoencoder.
In order to give the strangeness a global perspective, the exploration bonus is also augmented with the prediction error for the visited entire state.
We also build a new action-value function, exploration action-value function, on which a behavioral policy is formed.
We present how to train the function with the exploration bonus as well as the extrinsic reward, and make the original action-value function focus on training to make the goal policy optimal.
By using the proposed exploration method, the stochasticity of the multi-agent environment does not affect the exploration bonus largely.
We apply it to the well-known CTDE-based MARL algorithms and experimentally evaluate it on our didactic examples and the SMAC maps.
The results indicate that the proposed exploration method outperforms the existing methods and improves the MARL training performance significantly in the tasks requiring complex coordination.

%\bibliography{LinkPaper_2022_11}
%\bibliographystyle{IEEEtran}


\end{document}